\documentclass[10pt, a4paper]{article}

\usepackage{lrec-coling2024} 
 
\title{Clue-Instruct: Text-Based Clue Generation for Educational Crossword Puzzles}

\name{Andrea Zugarini$^1$, Kamyar Zeinalipour$^2$, Surya Sai Kadali$^2$,\\ {\bf \large Marco Maggini$^2$, Marco Gori$^2$, Leonardo Rigutini$^1$}}
\address{$^1$expert.ai, Siena, Italy\\ 
         $^2$University of Siena, Italy\\
         \{azugarini, lrigutini\}@expert.ai, suryasai.kadali@student.unisi.it,\\
         \{kamyar.zeinalipour2, marco.maggini, marco.gori\}@unisi.it
         }

\abstract{
Crossword puzzles are popular linguistic games often used as tools to engage students in learning. Educational crosswords are characterized by less cryptic and more factual clues that distinguish them from traditional crossword puzzles. Despite there exist several publicly available clue-answer pair databases for traditional crosswords, educational clue-answer pairs datasets are missing.
In this article, we propose a methodology to build educational clue generation datasets that can be used to instruct Large Language Models (LLMs). By gathering from Wikipedia pages informative content associated with relevant keywords, we use Large Language Models to automatically generate pedagogical clues related to the given input keyword and its context. 
With such an approach, we created \texttt{clue-instruct}, a dataset containing 44,075 unique examples with text-keyword pairs associated with three distinct crossword clues.
We used \texttt{clue-instruct} to instruct different LLMs to generate educational clues from a given input content and keyword. Both human and automatic evaluations confirmed the quality of the generated clues, thus validating the effectiveness of our approach.\\ \newline }

\makeatletter
\def\blfootnote{\xdef\@thefnmark{}\@footnotetext}
\makeatother

\begin{document}

\maketitleabstract

\blfootnote{Accepted for publication at LREC-COLING 2024, Turin, Italy, 20-25 May 2024.
}

\section{Introduction}\label{sec:inroduction}
\begin{figure*}[!t]
    \centering
    \includegraphics[scale=0.62]{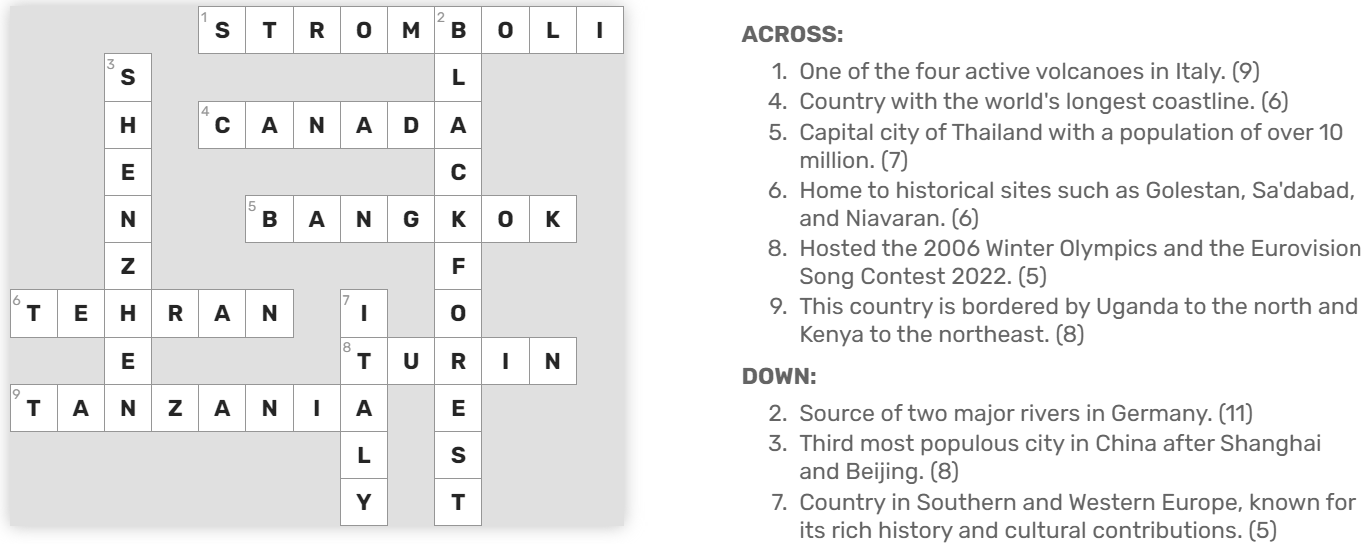}
    \caption{Example of an educational crossword puzzle on Geography-related keywords.}\label{fig:educw_example}
\end{figure*}

The conventional structure of crossword puzzles merged with scholastic elements results in an engaging learning tool: educational crosswords. They encompass a variety of subjects such as science, vocabulary, and history \cite{nickerson1977crossword, sandiuc2020use, yuriev2016crossword}. Educational crosswords differ from traditional puzzles because they are designed for teaching rather than entertainment. Consequently, their are less cryptic, and usually, they present a less constrained puzzle scheme, as in the example shown in Figure~\ref{fig:educw_example}. They are particularly beneficial in language acquisition or when mastering technical jargon for specific topics \cite{orawiwatnakul2013crossword, dzulfikri2016application, bella2023improving}. Additionally, the requirement of correlating appropriate hints with correct words fosters learners’ problem-solving skills \cite{kaynak2023effect, dol2017gpbl}. 
Memory enhancement is another merit of educational crosswords, as learners need to summon previously learned material to solve the puzzle \cite{mueller2018testing, dzulfikri2016application}. Moreover, the interactive nature of crosswords makes the learning experience captivating, inducing learners to persist in honing their abilities \cite{zirawaga2017gaming, bella2023improving}.
Summarily, educational crosswords serve as an entertaining resource for strengthening educational skills \cite{zamani2021use, yuriev2016crossword}.

Harnessing the power of Large Language Models (LLMs) presents an opportunity in the field of educational crossword production, traditionally known for requiring specialized skills and labor. 
Through an extensive training process on huge language corpora comprising internet resources, academic papers, and books, LLMs acquire the ability to generate high-quality text to accomplish many different tasks. This proficiency can be exploited to automatically generate clues, so to ease the process of educational crossword crafting.

In this work, we propose a methodology to construct datasets for educational crossword clue generation. In particular, we present \texttt{clue-instruct}, a corpus made of 44,075 clue generation instructions. Each example is constituted by a source text, serving as context, a category of interest, and a keyword, all paired with three target clues to generate. The dataset is built by gathering content from Wikipedia pages about relevant keywords, whereas clues were automatically generated by an LLM.
Upon \texttt{clue-instruct}, we carried out a detailed experimentation with different open-source LLMs varying in size and family, and we fine-tune them on the dataset. 
Results, assessed with both automatic and human evaluations, indicate that fine-tuning remarkably improves the generation quality of those models.
The dataset\footnote{\url{https://huggingface.co/datasets/azugarini/clue-instruct}} and all the models are publicly available.

The paper is organized as follows.
Section~\ref{sec:relatedworks} reports the related works on crosswords in NLP. We describe the proposed methodology in Section~\ref{sec:methodology} and analyse in detail the properties of the generated dataset in Section~\ref{sec:clueinstruct_dataset}. In Section~\ref{sec:experiments}, we discuss the experimental outcomes in-depth. Finally, we draw our conclusions in Section~\ref{sec:conclusions}.

\section{Related Works}\label{sec:relatedworks}

Crossword puzzles are a fascinating linguistic game that has been a subject of study in the Natural Language Processing field in the past few years. Literature can be divided into two main research branches: crossword solving and crossword generation~\cite{2010_Rigutini_LAP_Automatic_text_processing}. We briefly review both of them, then we finally discuss about existing crossword datasets. 

\paragraph{Crossword solving.}
Crossword resolution can be tackled as a constrained satisfaction task where the objective is to maximize the probability of filling the grid with answers coherent with the given clues. The main challenge in the problem is retrieving correct candidate answers.
Existing solutions heavily rely on clue-answer databases.
Proverb \cite{littman1999solving}, one of the earliest crossword-solving systems, used a probabilistic version of the A$^*$ with candidate answers retrieved from databases of American crosswords. 
Similarly, Dr. Fill~\cite{ginsberg2011dr} converted them into weighted CSPs and used advanced heuristics.
WebCrow \cite{ernandes2005webcrow, angelini2005webcrow, angelini2005solving} was a crossword-solving Italian project based on human-machine competitions. Webcrow distinguished from other solutions for exploiting the information present in the web. It was developed for the Italian language and English. Recently, it was extended to other languages~\cite{angelini2023webcrow,zugarini2023ratselrevolution} making use of neural representations of clue-answer pairs~\cite{zugarini2021multi}.
Based on WebCrow, SACRY leveraged syntactic structures for re-ranking and answer extraction to enhance answer quality by incorporating syntactic analysis~\cite{barlacchi2015sacry}.
Lately, the Berkeley Crossword Solver \cite{wallace2022automated} was presented. It was based on neural question-answering models for candidate answer retrieval, and belief propagation with local searches to fill the grid, achieving state-of-the-art performance in English crossword solving.

\begin{figure}[!t]
    \centering 
    \includegraphics[trim={1.2cm 1.3cm 0 1cm},clip,scale=0.52]{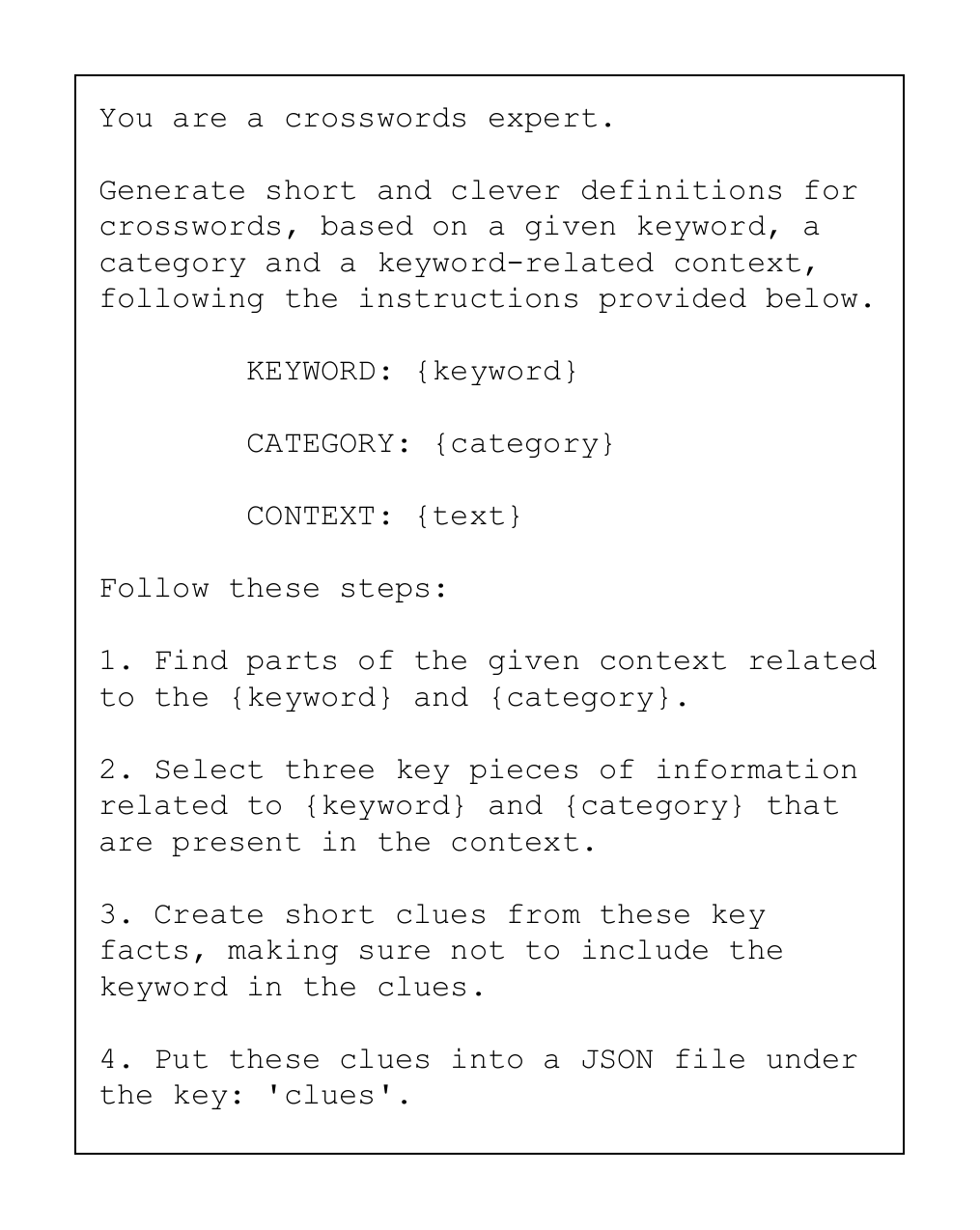}
    \caption{\texttt{clue-instruct} prompt used to generate the clues.}
    \label{fig:prompt}
\end{figure}

\begin{figure*}[ht]
    \centering 
    \includegraphics[trim={0 1.5cm 0 1.3cm},clip,scale=0.5]{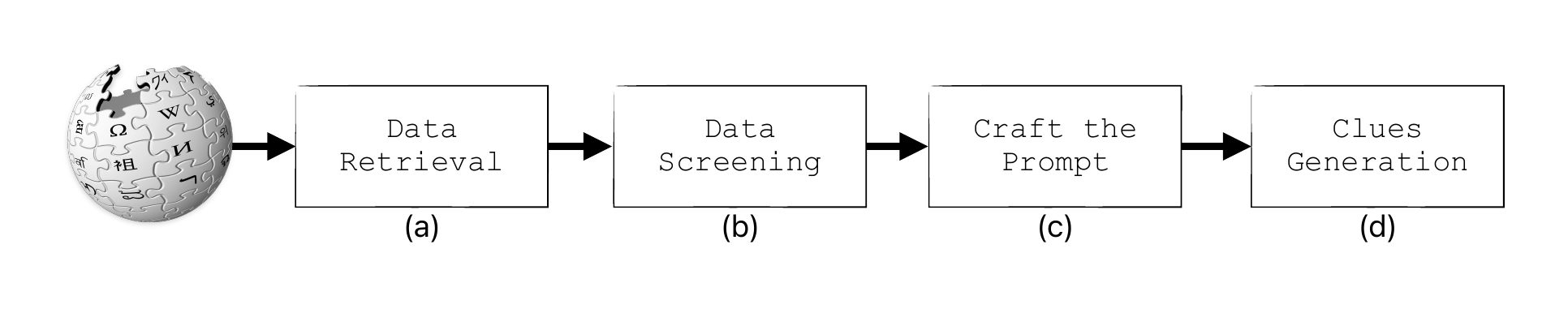}
    \caption{This figure illustrates the pipeline employed in constructing \texttt{clue-instruct}: (a) Information extraction from Wikipedia pages of text, keywords, and categories. (b) Data refinement and filtering to enhance data quality, by selecting the most crucial and highly-viewed pages, eliminating excessively short or overly detailed text, and more. (c) Design of the prompt for generating crossword clues based on input text and specified keywords within specific categories. (d) Exploit of GPT-3.5-Turbo to generate clues from the collected data and defined prompts.}
    \label{fig:dataset_method}
\end{figure*}

\paragraph{Crossword Generation.}
Building a crossword puzzle automatically encompasses different linguistic problems, such as identifying the answers, composing the grid, and above all, creating the clues. 
Early approaches \citet{rigutini2008fully, rigutini2012automatic} leveraged NLP techniques to generate lists of clue-answer pairs by analyzing online documents (Wikipedia pages). Clues were identified and extracted with NLP techniques (POS tagging, dependency analysis, and WordNet).
Analogously, the methods proposed in \citet{ranaivo2013automatic} and \citet{esteche2017automatic} followed a step-based approach to construct crosswords via NLP tools. The steps involved preliminary data extraction of sentences from a text that was then used to produce the clue-answer pairs.
A software tool utilizing NLP techniques to extract crucial keywords for crossword creation in Indian languages was proposed by \citet{arora2019automatic}. The resulting SEEKH framework combines statistical and linguistic methods to identify vital keywords.
More recently, \citet{10459980, zeinalipour-etal-2023-arabicros, zeinalipour2023italian} moved from handcrafted design of generated crosswords to generative solutions utilizing pre-trained LLMs. Crosswords were generated in English, Arabic, and Italian, thus demonstrating the effectiveness of computational linguistics in creating culturally diverse and engaging puzzles.
Analogously, our method makes use of LLMs to generate clues for a given answer, but we ground the generation to a source context with the purpose of producing clues that are adherent to a given input text.

\paragraph{Clue-Answer datasets.}
Despite many works have been published in both crossword solving and generation, few datasets have been created and publicly released. 
Most of them consist of clue-answer pair corpora, generally collected from crosswords or clue databases~\cite{Ernandes2008Challenges, ginsberg2011dr} sometimes enriched by metadata such as publication date, publisher, and difficulty. Unfortunately, for copyright reasons, they are not always publicly available.
In \cite{barlacchi2015sacry}, to test their proposed system, the authors created a corpus by downloading crossword puzzles from some web sources.
\citet{wallace2022automated} collected a validation and test set of complete 2020 and 2021 puzzle grids from several US news (The New York Times, The LA Times, Newsday, The New Yorker, and The Atlantic) and they publicly released code, models, and dataset.
However, all these clue-answer pairs corpora are constructed from traditional crossword puzzles. In these types of puzzles, the clues usually have extremely enigmatic linguistic structures that are quite different from those typically adopted for educational purposes.
Furthermore, by design they lack of any reference to textual passages in which the clue can be found. This information is very important in the educational use-case where the clue must be related to a subject of study. Moreover, a grounding context allow to steer the generation of a Language Model, thus dramatically reducing the occurrence of hallucinated or unrelated clues. 

In this work instead, we propose a method to create a clue generation corpus where clues are tied with an answer and a source context. The obtained dataset is, to the best of our knowledge, the first corpus associating such information together.

\section{Method}\label{sec:methodology}
Differently from traditional clue-answer crossword databases, we necessitate aligning the clue-answer pair with a grounding text, where the answer to the clue can be inferred from it. 
The grounding text is crucial in education both from the perspective of a teacher and from the point of view of the student.
In order to construct such a context-keyword-clue triplet, we follow a pipeline, starting from collecting and gathering data from Wikipedia. The entire pipeline is sketched in Figure~\ref{fig:dataset_method}. Here we describe it step by step.

\paragraph{Data Retrieval.} \label{sec:datret}
We initiate the information extraction process by mining Wikipedia pages. This involves accessing the initial section of each page, which typically contains the most pertinent information. From this portion, we emphasize keywords presented in bold, which often correspond to the page's title but can include additional terms. These selected keywords become the focal points of the Wikipedia page, shaping the content to provide in-depth definitions and explanations. In addition to the content, we gather various metadata about the page, including the number of page views, an overall importance rating, text within paragraphs, its title, associated keywords, relevant categories, and individual URLs. Leveraging the standardized layout of Wikipedia pages, we extract keyword-rich opening paragraphs that encapsulate the core content, offering succinct explanations or definitions, and contributing to the construction of a valuable dataset.

\begin{figure}[t]
    \centering 
    \includegraphics[trim={0 0 0 0.4cm},clip,scale=0.8]{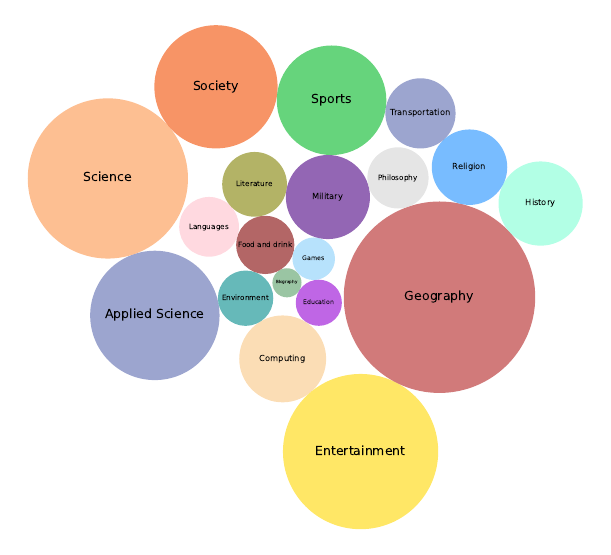}
    \caption{Distribution of the examples among the twenty categories.}
    \label{fig:topics-distr}
\end{figure}

\paragraph{Data Screening.} \label{sec:datscr}
With the goal of discarding low-quality data, we adopt several filters: (1) We select pages based on the number of views and importance rating; (2) We remove pages with too long or too short contents; (3) All the data with keywords made of more than three words were removed; (4) all the keywords outside typical English crossword boundaries -- $[3,20]$ character length -- or containing non-alphabetical symbols were excluded.
\paragraph{Craft the prompt.} \label{sec:datprompt}
The creation of an effective prompt was a crucial aspect of our methodology. We carefully designed prompts for crossword clue generation by incorporating the relevant keywords extracted from the Wikipedia pages. These prompts were structured to provide contextual guidance for generating clues that were both informative and engaging. By using the extracted keywords along with the context of the Wikipedia page, the prompts acted as input signals to guide the generation of crossword clues. Our goal was to create prompts that were well-suited to each specific topic or subject area, taking into account the unique characteristics of the information we had gathered. Crafting the prompt effectively played a key role in the success of our approach, enabling our system to produce high-quality crossword clues tailored to educational needs. In Figure \ref{fig:prompt}, the prompt employed in the study is depicted.

\paragraph{Clues Generation.} \label{sec:datagen}
After assembling content, keywords and categories into the prompt, in the last pipeline step, we generate educational clues for such data. Inspired by \textsc{self-instruct}~\citep{wang2022self}, we make use of Large Language Models for automatically generating clues. Differently from \textsc{self-instruct}, generation is strongly conditioned by the information in the input context of the LLM. Therefore, we expect it to produce more faithful clues, thus significantly mitigating the risks of hallucinations.  

\begin{table}[t]
    \centering
    \begin{tabular}{c c}
    \hline
    \multicolumn{2}{c}{\textbf{clue-instruct}}\\
    \hline
         \# contexts & 44,075\\
         \# keywords & 44,075\\ 
         \# categories & 20\\
         \# clues & 132,225\\
         \hline
    \end{tabular}
    \caption{General statistics on \texttt{clue-instruct} dataset.}
    \label{tab:stat}
\end{table}

\begin{figure}[ht]
    \centering
    \includegraphics[scale=0.32]{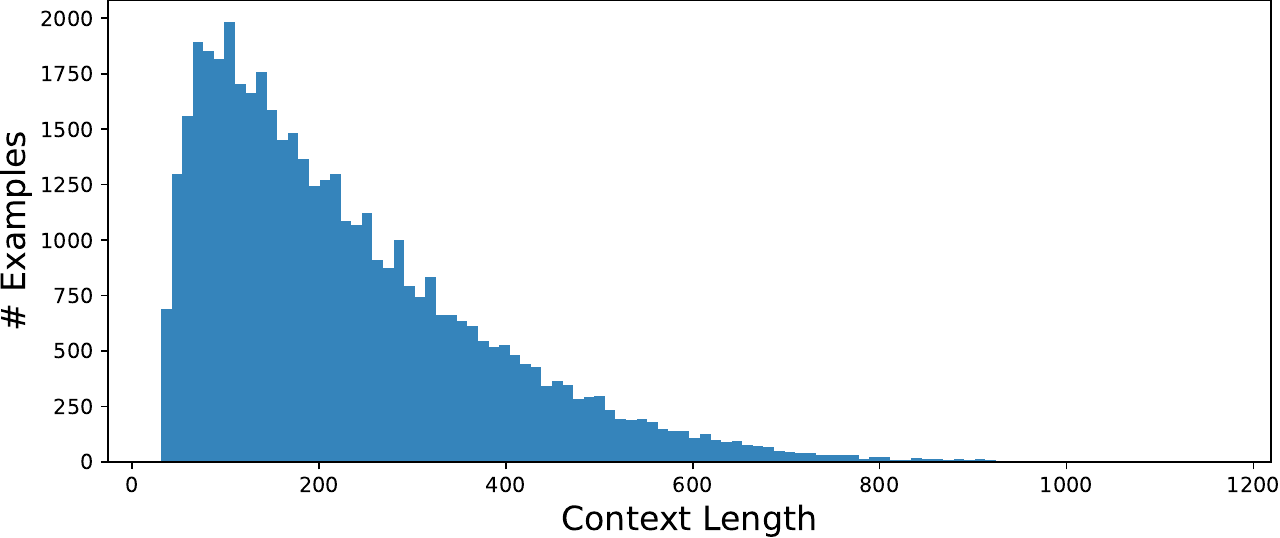}\\
    \includegraphics[scale=0.32]{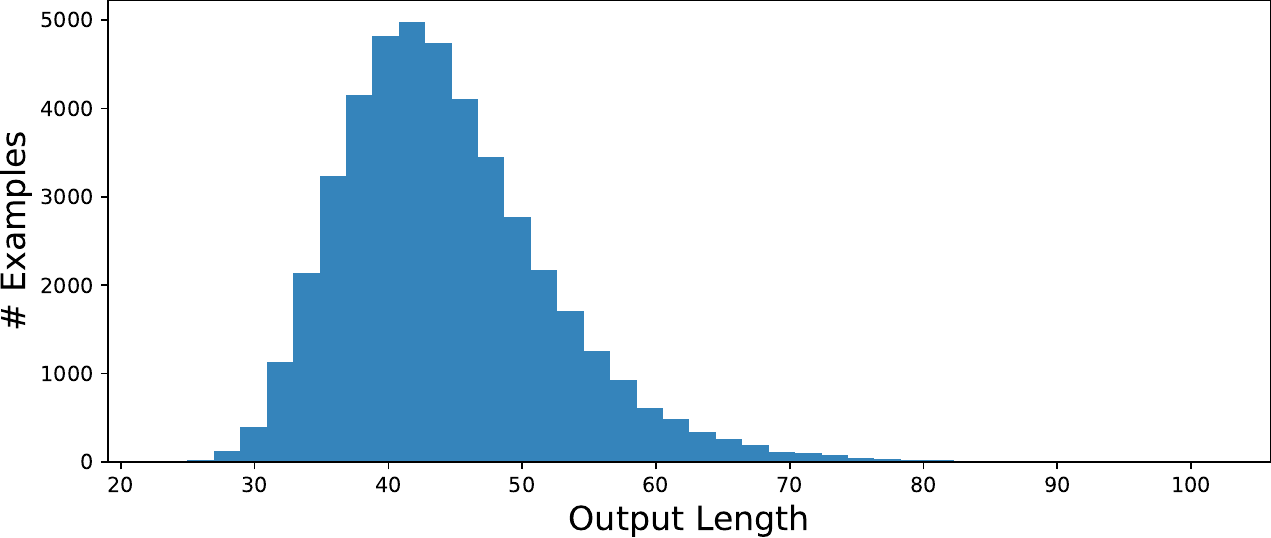}\\
     \includegraphics[scale=0.32]{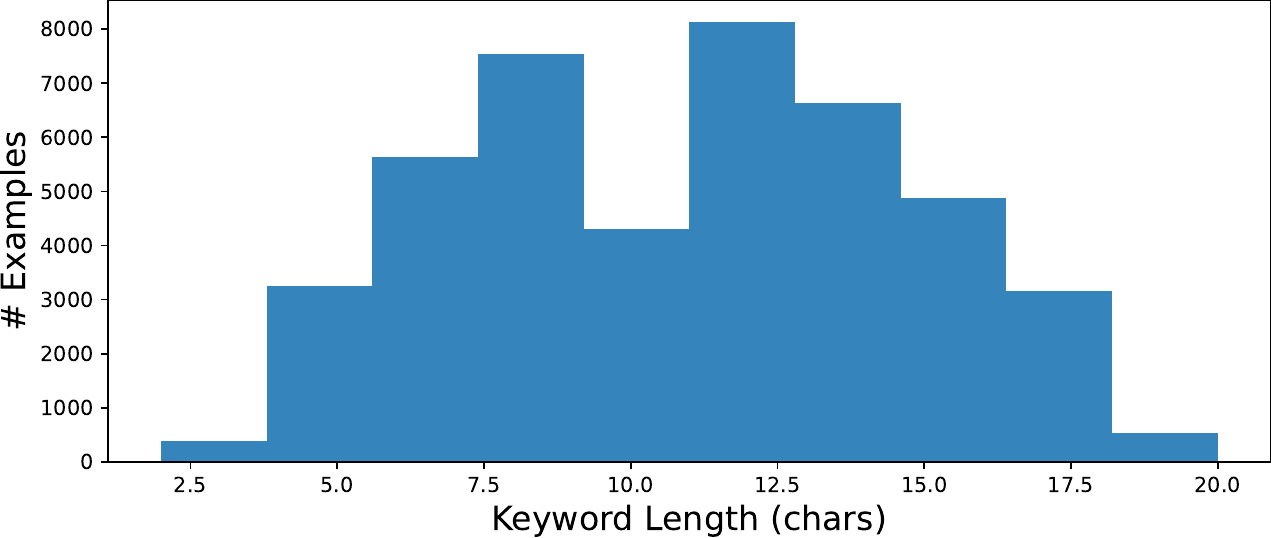}
    \caption{Word length distribution of contexts and outputs. Char length distribution over keywords. }
    \label{fig:dataset_distrubtions}
\end{figure}

\begin{figure}
    \centering \includegraphics[scale=0.35]{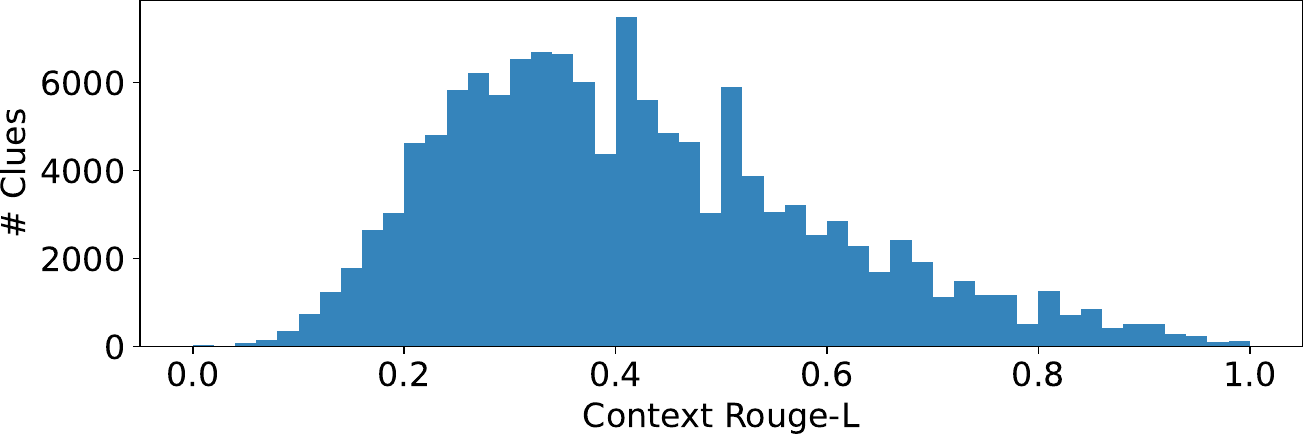}
    \caption{ROUGE-L score distribution of clues vs sentences in wiki page contents computed over the entire \texttt{clue-instruct} dataset.}
    \label{fig:context_rouge_full_data}
\end{figure}

\begin{table*}[]
    \centering
    \begin{tabular}{ccp{0.4\textwidth}c}
\textbf{Answer}&\textbf{Category}&\centering\textbf{Clue}&\textbf{Rating}\\ \hline
Robocall & Society & May be blocked by phone companies to prevent scams & A \\
Ministry Of Magic & Literature & Corrupt and incompetent government in J.K. Rowling's Wizarding World & A \\
Lovesick & Literature & Renewed for a third season, released exclusively on Netflix & C \\
South American tapir & Science & One of the four recognized species in the tapir family & E \\ 
\hline
    \end{tabular}
    \caption{Some examples of generated clues alongside the human rating assigned to each of them.}
    \label{tab:cluegen_examples}
\end{table*}

\begin{figure}
    \centering
    \includegraphics[scale=0.33]{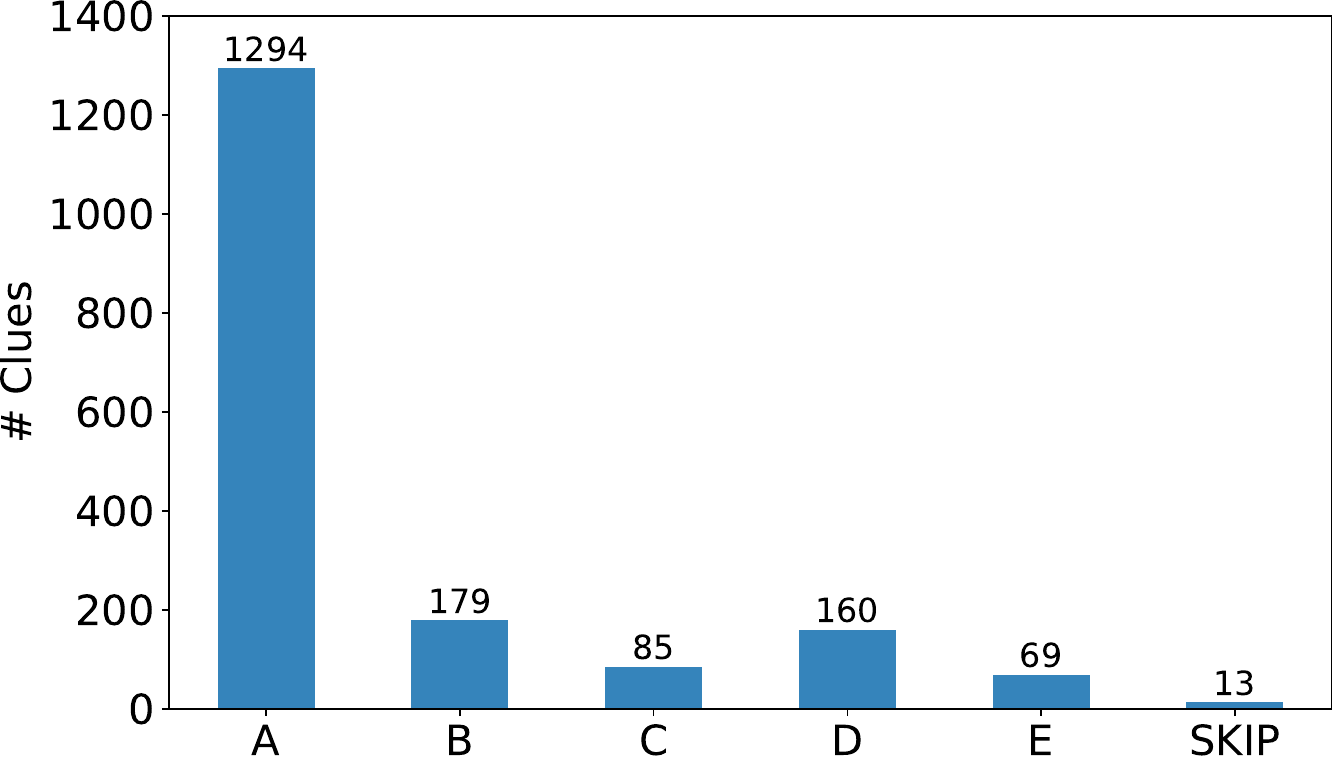}
    \caption{Ratings assigned by humans on the test set.}
    \label{fig:ratings_plot}
\end{figure}

\section{Clue-Instruct Dataset}\label{sec:clueinstruct_dataset}
With the method described in Section~\ref{sec:methodology}, we construct a dataset for English educational clues, starting from the most popular pages of 20 distinct categories, that initially contained 258,325 Wikipedia pages. 
We kept all the pages with more than 10,000 views or with an importance rating equal to 'Top'. Contexts below 30 or above 1000 words were deleted. After data screening, we obtained a corpus of 44,075 examples in total. We used GPT-3.5 Turbo~\citep{brown2020language} as clue generator LLM, with the prompt depicted in Figure~\ref{fig:prompt}. 

\subsection{Statistics}
In this section, we delve into the statistical properties of \texttt{clue-instruct}.
Table \ref{tab:stat} presents an overview of the dataset. It comprises overall 44,075 textual content-keyword pairs across 20 different categories. Three distinct clues were generated from each content-keyword-category triplet, resulting in a total of 132,225 clues.

As previously highlighted, examples are divided into 20 distinct categories. In Figure~\ref{fig:topics-distr}, we visually represent the frequency of each category within our dataset using a bubble plot. The size of each bubble corresponds to the frequency of the respective category in the dataset. Upon analyzing this plot, it becomes evident that 'Geography', 'Science', and 'Applied Science' are the most prevalent categories, in that order. Conversely, 'Biography', 'Games', and 'Education' are the least frequent ones.

In Figure~\ref{fig:dataset_distrubtions}, we outline the distribution of context and output lengths in relation to the number of words. Such a figure also presents the keyword length distribution in terms of characters. 
We can observe how the context length falls in a wide range going from 30 to 1000 words, with the vast majority of examples having context lengths between 50 to 400. Conversely, most outputs have word lengths between 35 to 50. 
Additionally, the keyword character length spans from 3 to 20 as imposed during the corpus creation. 

\subsection{Measuring Data Quality}
To evaluate the dataset quality, we resorted to both automatic metrics and human evaluations.

\paragraph{Automatic Metrics.} Due to the absence of a reference corpus for educational crosswords, there is no reference set to compare the generated clues with. Therefore, we cannot produce standard automatic metrics such as ROGUE scores.
Nonetheless, in the specific educational clue generation task, good clues should tightly adhere to the reference context, being simple reformulations of some information stated in the text. Hence, the problem is highly extractive. 

From such considerations, we exploited as automatic evaluation, the ROUGE-L score between the sentences in the input context against the generated clue. 
Intuitively, scores should be high enough to indicate strong adherence to the context, thus reducing the chances of hallucinations, but not too close to perfect matches, which would be an indication of poor clue styling and high chances of injecting the target keyword within the clue itself. 
On average, we obtained about 42 ROUGE-L, which indicates a significant entailment between the generated clue and the most similar sentence in the context. The distribution over the dataset is outlined in Figure~\ref{fig:context_rouge_full_data}.

\begin{table*}[!ht]
     \centering
     \begin{tabular}{cccccc}
      \hline
    \textbf{model type}&    \textbf{model name} & \textbf{\# params} & \textbf{ROUGE-1}& \textbf{ROUGE-2} & \textbf{ROUGE-L} \\ \hline
     &\textsc{Llama2-chat}  & 7B & -- & -- & --\\
   Off-the-shelf LLMs &\textsc{mpt-instruct}  & 7B & 23.98 & 11.79 & 19.69\\
    &\textsc{Llama2-chat}  & 13B & \textbf{31.80} & \textbf{15.32} & \textbf{25.27}\\
    &\textsc{mpt-instruct}  & 30B & 29.92 & 14.47 & 24.30\\
     \hline
   
    &\textsc{Llama2-chat}  & 7B & 59.92 & 40.98 & 52.28\\
   Finetuned LLMs &\textsc{mpt-instruct}  & 7B & 59.26 & 40.37 & 51.68\\
    &\textsc{Llama2-chat}  & 13B & \textbf{62.97} & \textbf{44.97} & \textbf{55.40}\\
    &\textsc{mpt-instruct}  & 30B & 61.42 & 42.63 & 53.77\\   \hline
     \end{tabular}
     \caption{Performance of off-the-shelf LLMs with and without fine-tuning. Without clue generation instruction tuning, smaller models struggle to follow the request. Fine-tuning greatly improves the performances of all the LLMs.}
     \label{tab:llms_results}
 \end{table*}

\paragraph{Human Evaluation.}
To assess the quality of generated data, we cannot solely rely on automatic metrics. Thus, we sample a portion of clue-instruct for human evaluation. Similarly to \citet{wang2022self}, we consider a five-level rating, under the following guidelines:
\begin{itemize}
    \item \texttt{RATING-A}: The clue is valid and coherent to the given context, answer, and category.
    \item \texttt{RATING-B}: Acceptable clue with minor imperfections - loose correlation with category.
    \item \texttt{RATING-C}: The clue is relevant to the answer but loosely correlates with the context, or it is too generic.
    \item \texttt{RATING-D}: The clue is irrelevant and/or incorrect with respect to the answer or the context.
    \item \texttt{RATING-E}: Not acceptable clue because it contains the answer (or a variant of it).
\end{itemize}
We also allow annotators to skip examples (marked with \texttt{SKIP}), in case there are issues not strictly related to the clue itself, such as odd keywords or documents. 

Overall, 600 examples were annotated, for a total of 1,800 clues evaluated, since there are three clues proposed by the model for each given context, keyword, and category triplet. We report rating distributions in Figure~\ref{fig:ratings_plot}. More than two out of three  (about 72\%) clues were marked with \texttt{RATING-A}, the highest score, which grows to 81\% if we consider as acceptable also the clues rated with \texttt{B}.

\begin{figure}[h]
    \centering 
    \includegraphics[scale=0.31]{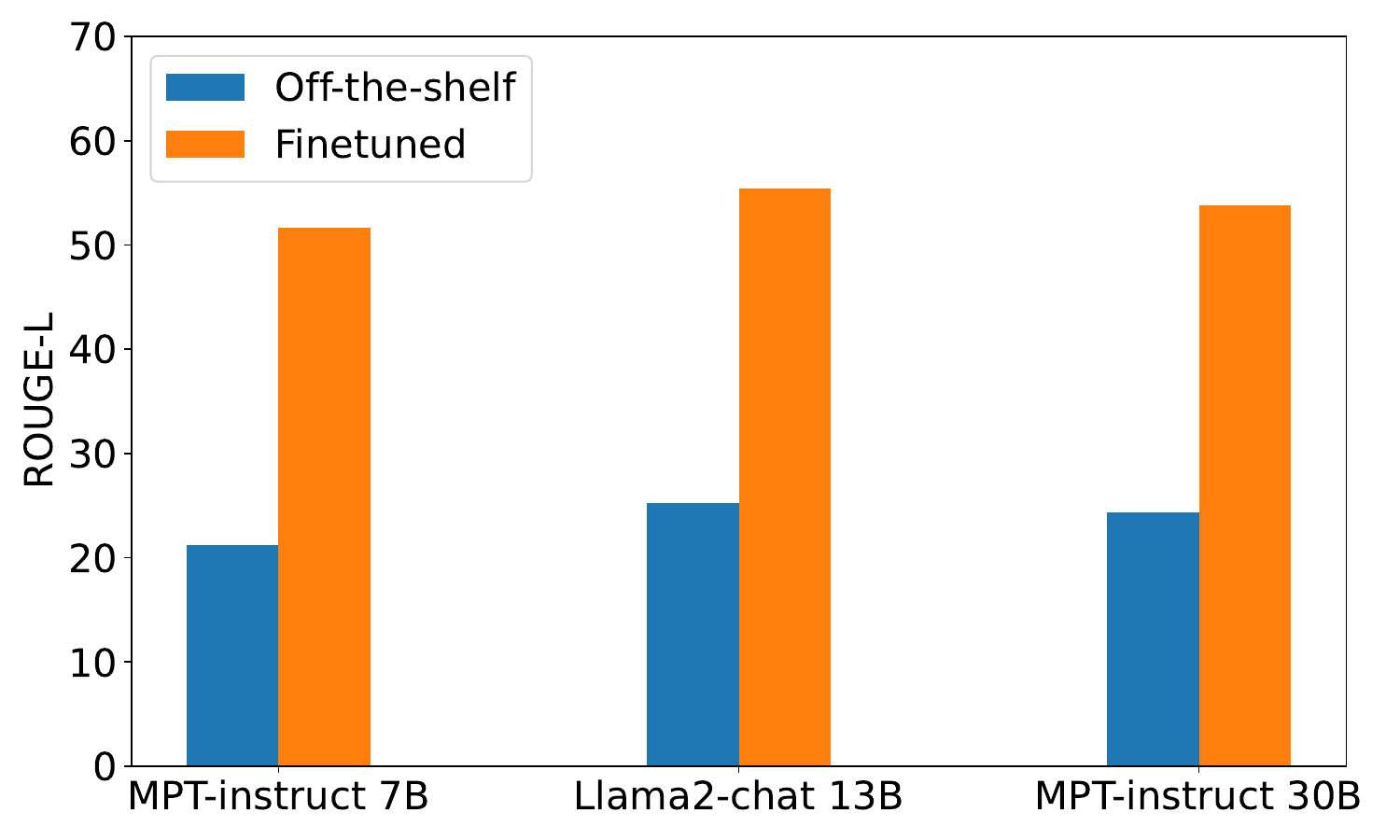}
    \caption{ROUGE-L of different LLMs with and without finetuning.}
    \label{fig:cmp_finetuned_nonfinetuned_llms}
\end{figure}

 \begin{figure*}[ht]
    \centering 
    \includegraphics[scale=0.46]{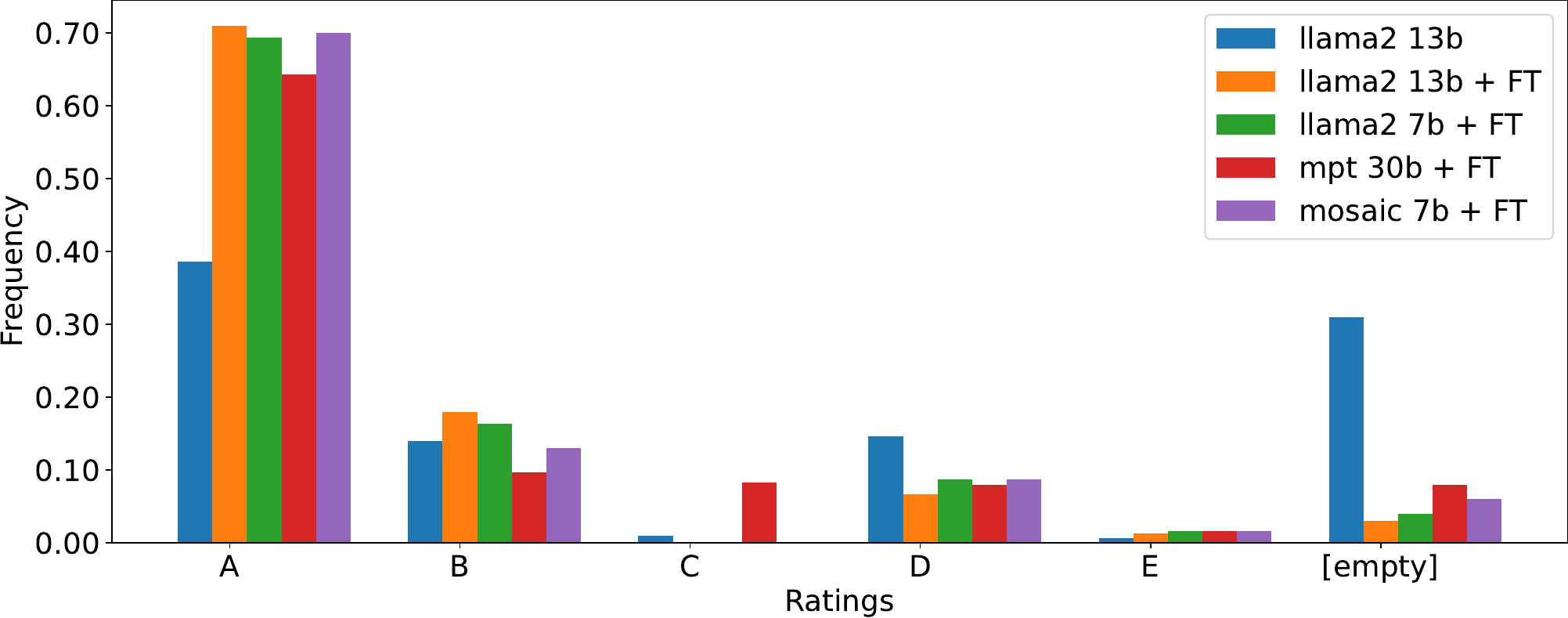}
    \caption{Human evaluation of clues generated by different LLMs. Models finetuned on \texttt{clue-instruct} are indicated with ``+ FT''.}
    \label{fig:human-eval_llms}
\end{figure*}
 
\section{Experiments}\label{sec:experiments}
We investigate the usage of \texttt{clue-instruct} to fine-tune different families of LLMs with various dimensions. 

\subsection{Experimental Setup} 
\paragraph{Data.} LLMs were trained for instruction tuning on \texttt{clue-instruct}. We kept the 600 annotated examples as a test and used them to evaluate all of our models using GPT-3.5 Turbo as an oracle. The remaining 43,475 examples were used for training. 
LLMs were instructed with the same prompt used for GPT-3.5 Turbo depicted in Figure~\ref{fig:prompt}. 

\paragraph{Baselines.} We focus on four instruction-tuned LLMs: \textsc{Llama2-chat}~\citet{touvron2023llama} in 7B and 13B sizes, and \textsc{\textsc{mpt-instruct}}~\citet{MosaicML2023Introducing} in both 7B and 30B releases. 

\paragraph{Training details.}
All the models were fine-tuned with LORA~\cite{hu2021lora}, $r=16$, and $\alpha=32$ over the course of two training epochs and batch size set to 32. Learning rate was initialized to $3\cdot 10^{-4}$ with a linear warm-up of 200 steps.
At inference time, clues were generated by sampling from the model distribution. The temperature was set to 0.1, while top-$p$ and top-$k$~\citep{holtzman2019curious} were set to 0.75 and 50, respectively. All the experimentation was carried out on a server equipped with four NVIDIA A6000 GPUs. 

\subsection{Results}
\paragraph{Off-the-shelf LLMs.}
First of all, we evaluate the four baseline models in zero-shot, i.e. without any fine-tuning on \texttt{clue-instruct}. Comparison is shown in Table~\ref{tab:llms_results}. Despite being previously trained to follow generic instructions, all the models struggle to produce a valid set of clues. Results are in general not satisfactory. In particular, \textsc{Llama2}7B{-chat} always fails to produce an acceptable output with the given prompt. Probably, different prompt designs would have led to better results for such a model, however, this inquiry goes beyond the goals of our paper. Also \textsc{mpt-instruct}-7B often poorly fails to produce the correct JSON output and often generates a single clue, instead of the three requested. With the increase of models' parameters, also the quality grows. Both \textsc{Llama2-}13B\textsc{-chat} and \textsc{mpt-instruct}-30B have higher ROUGE-L scores, with the former slightly better than the latter. This is mainly due to the fact that \textsc{Llama2-}13B\textsc{-chat} always produced the exact JSON schema, whereas \textsc{mpt-instruct-}30B failed almost once every four times.

\paragraph{Finetuned LLMs.} When finetuning the baseline LLMs on \texttt{clue-instruct}, all the models exhibit a remarkable improvement. Such a comparison is clearly shown in Figure~\ref{fig:cmp_finetuned_nonfinetuned_llms}. ROUGE-L results are outlined in Table~\ref{tab:llms_results}. The outputs always align with the expected format. Finetuned LLMs surpass off-the-shelf models by a large margin, with an increase above 20 points in ROUGE-L. It is worth noticing that, \textsc{Llama2-chat} 13B is confirmed to be the best model, and that \textsc{Llama2-chat} 7B can recover from catastrophic results.
All the finetuned LLMs are publicly available\footnote{\url{https://huggingface.co/azugarini/clue-instruct-llama-7b}}\textsuperscript{,}\footnote{\url{https://huggingface.co/azugarini/clue-instruct-llama-13b}}\textsuperscript{,}\footnote{\url{https://huggingface.co/azugarini/clue-instruct-mpt-7b}}\textsuperscript{,}\footnote{\url{https://huggingface.co/azugarini/clue-instruct-mpt-30b}}. 

\paragraph{Impact of model size and LLM family.}  Analyzing the results from Table~\ref{tab:llms_results}, we can notice that larger models tend to outperform smaller ones. In particular, larger LLMs are more robust to unseen instructions, thus showing wider gaps when not finetuned on the downstream task. Moreover, we can observe that \textsc{Llama2-chat} 13B model is particularly well-performing, surpassing \textsc{mpt-instruct} 30B, which is more than twice its size, as already observed in the literature.

\begin{figure}[!hb]
    \centering
    \includegraphics[scale=0.355]{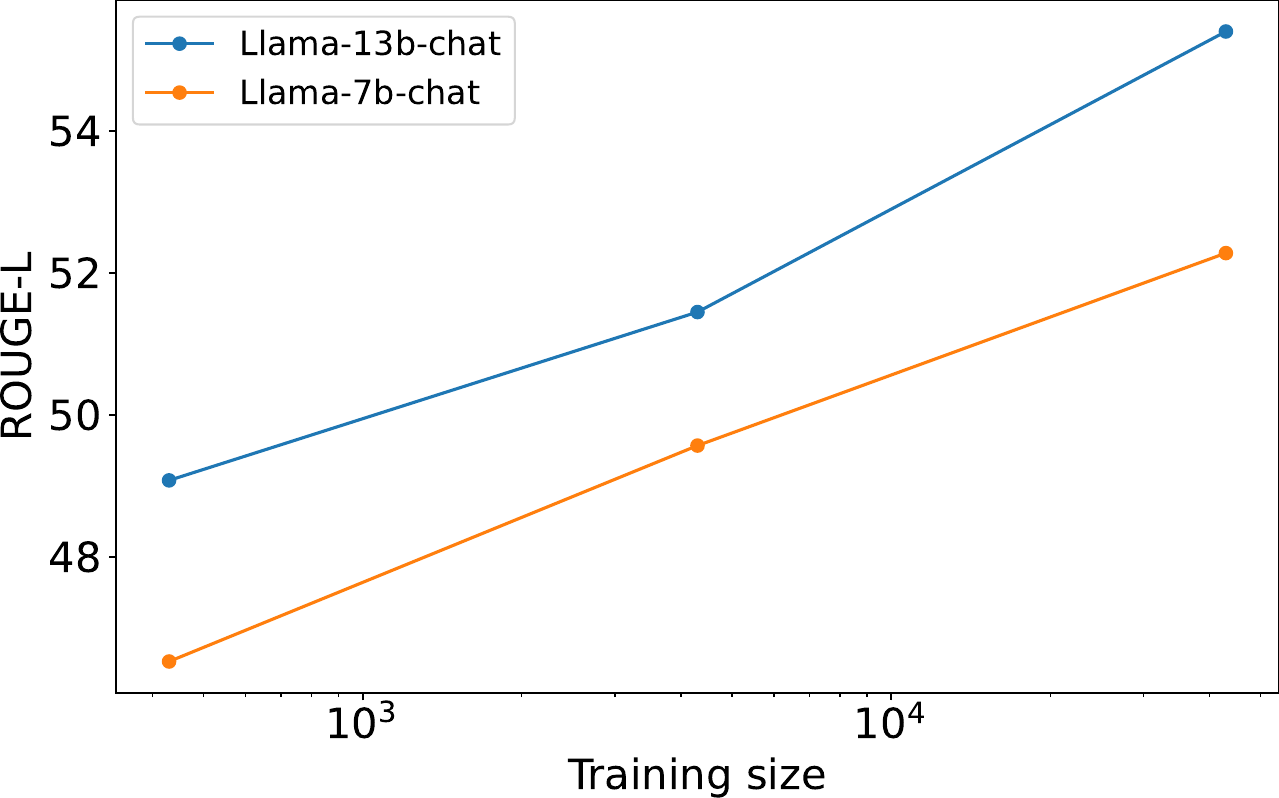}
    \caption{Impact of training size on ROUGE-L. \texttt{clue-instruct} is truncated at sizes corresponding to $1\%$,$10\%$ and $100\%$ of the training corpus.}
    \label{fig:train_size_trend}
\end{figure}

\paragraph{Impact of dataset size.} 
We also measure how the performance changes when using different amounts of training examples. Training size was cut at $1\%$, $10\%$, and $100\%$, to see the trend at different orders of magnitude. To slightly cope with the reduced amount of training steps, we increase the number of epochs to 3 for $1\%$ and $10\%$ pieces of training, and we reduce the number of warm-up gradient steps to $20$. In this experiment, we only focus for simplicity on the \textsc{Llama2} family (7B, 13B). From the results, outlined in Figure~\ref{fig:train_size_trend}, we can observe that a small number of examples are enough to align the LLMs to the task, even for \textsc{Llama2-chat}-7B that failed to produce valid clues when applied as zero-shot. Thus, the biggest leap in performance is given by just a small amount of instructions, coherently with findings in literature \citep{zhou2023lima}.  

\paragraph{Human Evaluation.} In addition to automatic evaluation, human annotators are asked to evaluate the output of the fine-tuned LLMs. We compare the models on a portion of 100 documents of the test set. Due to the poor performance of off-the-shelf models, we only consider \textsc{Llama2-chat}-13B in this evaluation with the purpose of highlighting once again the differences between base models and their fine-tuned versions on \texttt{clue-instruct}. In addition to rating scores, we marked as \textsc{[EMPTY]} all examples where a clue was not produced. We report the results in Figure~\ref{fig:human-eval_llms}. All the tuned models exhibit a major reduction of malformed outputs (\textsc{[EMPTY]}). In contrast, the number of A-rated examples suddenly increased.
Also, D-rated examples diminish, whereas B, C, and E rates have a slight increase, with some exceptions. These results suggest that finetuning is extremely effective in aligning the generated output to the expected format, but there is also a positive contribution to the quality of the generated clues. 
To help understanding what kind of clues were generated and the ratings assigned, we showcase some examples in Table~\ref{tab:cluegen_examples}.

\section{Conclusions}\label{sec:conclusions}
In this paper, we presented a methodology to generate clues for educational crosswords, from which we constructed \texttt{clue-instruct}, an instruction-tuning dataset with keyword-clue pairs grounded on an input context, specifically designed for educational crosswords. 
To the best of our knowledge, the corpus is the first resource that combines such information, which is necessary to build systems that can generate educational crosswords from a given document. 
We then leveraged \texttt{clue-instruct} to fine-tune different open-source Large Language Models, showing that aligning LLMs to this kind of instructions greatly improves the output quality in terms of both automatic and human evaluation. Both the dataset and the models have been publicly released.

In the future, we plan to further extend our methodology to non-English languages in order to facilitate the diffusion of educational crosswords also in less represented languages.

\section{Acknowledgements}
This work was supported by the IBRIDAI
project financed by the Regional Operational Program ``FESR 2014-2020'' of Emilia
Romagna (Italy), resolution of the Regional Council n. 863/2021.

\nocite{*}
\section{Bibliographical References}\label{sec:reference}

\bibliographystyle{lrec-coling2024-natbib}
\bibliography{lrec-coling2024}

\end{document}